\documentclass[sigconf]{acmart}
\usepackage{listings}

\AtBeginDocument{%
  \providecommand\BibTeX{{%
    \normalfont B\kern-0.5em{\scshape i\kern-0.25em b}\kern-0.8em\TeX}}}

\setcopyright{acmcopyright}
\copyrightyear{2020}
\acmYear{2020}

\acmConference[KDD '20]{KDD '20: ACM SIGKDD Conference on Knowledge Discovery and Data Mining}{August 22--27, 2020}{San Diego, CA}
\acmBooktitle{KDD '20: ACM SIGKDD Conference on Knowledge Discovery and Data Mining,
  August 22--27, 2020, San Diego, CA}

\usepackage{multirow}
\usepackage{flushend}
\usepackage{caption}
\begin{document}

\title{Map Generation from Large Scale Incomplete and Inaccurate Data Labels}

\author{Rui Zhang}
\email{rui.zhang@ibm.com}
\affiliation{%
  \institution{IBM T.J. Watson Research Center}
  \streetaddress{1103 Kitchawan Rd.}
  \city{Yorktown Heights}
  \state{New York}
  \postcode{10598}
}

\author{Conrad Albrecht}
\email{cmalbrec@us.ibm.com}
\affiliation{%
  \institution{IBM T.J. Watson Research Center}
  \streetaddress{1103 Kitchawan Rd.}
  \city{Yorktown Heights}
  \state{New York}
  \postcode{10598}
}

\author{Wei Zhang}
\email{weiz@us.ibm.com}
\affiliation{%
  \institution{IBM T.J. Watson Research Center}
  \streetaddress{1103 Kitchawan Rd.}
  \city{Yorktown Heights}
  \state{New York}
  \postcode{10598}
}

\author{Xiaodong Cui}
\email{cuix@us.ibm.com}
\affiliation{%
  \institution{IBM T.J. Watson Research Center}
  \streetaddress{1103 Kitchawan Rd.}
  \city{Yorktown Heights}
  \state{New York}
  \postcode{10598}
}

\author{Ulrich Finkler}
\email{ufinkler@us.ibm.com}
\affiliation{%
  \institution{IBM T.J. Watson Research Center}
  \streetaddress{1103 Kitchawan Rd.}
  \city{Yorktown Heights}
  \state{New York}
  \postcode{10598}
}

\author{David Kung}
\email{kung@us.ibm.com}
\affiliation{%
  \institution{IBM T.J. Watson Research Center}
  \streetaddress{1103 Kitchawan Rd.}
  \city{Yorktown Heights}
  \state{New York}
  \postcode{10598}
}

\author{Siyuan Lu}
\email{lus@us.ibm.com}
\affiliation{%
  \institution{IBM T.J. Watson Research Center}
  \streetaddress{1103 Kitchawan Rd.}
  \city{Yorktown Heights}
  \state{New York}
  \postcode{10598}
}

\renewcommand{\shortauthors}{Zhang, et al.}

\begin{abstract}

Accurately and globally mapping human infrastructure is an important and challenging
task with applications in routing, regulation compliance monitoring,
and natural disaster response management etc.. In this paper we present progress
in developing an algorithmic pipeline and distributed compute system that
automates the process of map creation using high resolution aerial images.
Unlike previous studies, most of which use datasets that are available only in a
few cities across the world, we utilizes publicly available imagery and map data,
both of which cover the contiguous United States (CONUS).
We approach the technical challenge of inaccurate and incomplete training data adopting
state-of-the-art convolutional neural network architectures such as the U-Net and
the CycleGAN to incrementally generate maps with increasingly more accurate and more
complete labels of man-made infrastructure such as roads and houses.
Since scaling the mapping task to CONUS calls for parallelization, we then adopted
an asynchronous distributed stochastic parallel gradient descent training scheme to distribute
the computational workload onto a cluster of GPUs with nearly linear speed-up.
\end{abstract}



%


\begin{teaserfigure}
\centering
\captionsetup{justification=centering}
\vspace{-0.5cm}
  \includegraphics[width=.8\textwidth]{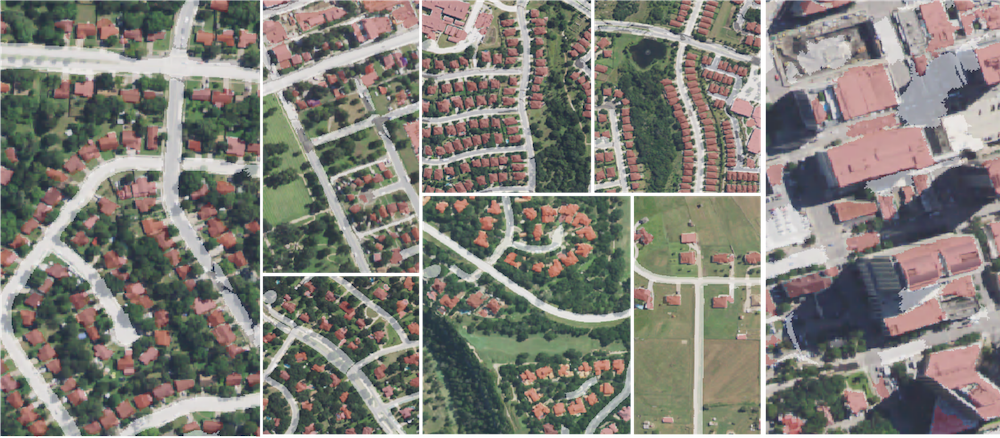}
  \caption{%
    Houses (red semi-transparent boxes) and roads (white semi-transparent lines)\newline
    generated from aerial imagery (background).
  }
    \vspace{-0.1cm}
  \Description{}
  \label{fig:teaser}
\end{teaserfigure}


\maketitle

\section{Introduction}
\label{sec:intro}
Generating maps of roads and houses from high resolution imagery is critical to
keep track of our ever changing planet. Furthermore, such capability become vital
in disaster response scenarios where pre-existing maps are often rendered useless
after destructive forces struck man-made infrastructure. New maps of roads accessible
as well as indication of destroyed buildings will greatly help the disaster response
team in rescue planning.

Due to the significant advancement in computer vision by deep learning during the
last couple of years in parallel to the explosive amount of high-resolution imagery
becoming available, there has been a growing interest in pushing research in automated
map generation from high resolution remote sensing imagery
\cite{spacenet,RefineNetMicrosoft2018,kaiser2017learning}.
Several challenges in the field of detecting houses or
roads using high-resolution remote sensing data have been established--two of them
being SpaceNet \cite{spacenet} and DeepGloble \cite{DeepGlobe18}. SpaceNet hosted
a series of challenges including three rounds of building detection competitions,
and two rounds of road network detection contests.
The SpaceNet building detection challenge provides 5 cities (plus one additional
city--namely Atlanta, GA, USA--in the last round) across different continents,
including Las Vegas, NV, USA; Paris, France; Rio de Janeiro, Brasil; Shanghai, China;
and Khartoum, Sudan. The corresponding building and house labels are provided along
with the high resolution imagery. The DeepGlobe challenge was using the same set
of data, but employed a slightly different evaluation measure.

The SpaceNet and DeepGlobe training datasets being publicly available attracted
researchers globally to address the challenge of automated map generation.
However, as is commonly known infrastructure vary significantly from one
location to another. While high-rise apartment buildings have become the typical
in China whereas most US population lives in single family houses. A model trained
on US data will perform poorly on China. Even within the US from state to state
the geo-spatial variation is evident.
From this perspective a task of training a model to generate a global map requires
more geographically diverse training data. Indeed, the past decade has shown a
dramatic increase in the amount of open geo-spatial datasets made available by
government agencies such as the
\href{https://www.usda.gov/}{United States Department of Agriculture} (USDA), \href{https://www.nasa.gov/}{NASA},
the \href{https://www.esa.int/}{European Space Agency}. Specifically in this paper,
we employ two publicly available datasets to train the model of map generation from imagery.
One from the corpus of OpenStreetMap (OSM) data and another from the National Agriculture
Imagery Product (NAIP) distributed by USDA.


Most state-of-the-art automated programs rely on a significant amount of high-resolution
image data with a correspondingly well labeled map
\cite{kaiser2017learning,maggiori2016convolutional}. There has been limited studies
utilizing the freely available OSM due to its fluctuating accuracy depending on
the OSM community's activity in a given geo-location.
In this work we attempt to utilize the inaccurate and incomplete labels of OSM
to train neural network models to detect map features beyond OSM.

In addition to the challenge in data quality, the data volume needed to train a
model that can map the whole United States is daunting too. The total
area of the CONUS is about $7.7$ million square kilometers. For NAIP imagery of
$1~m$ resolution, the amount of data sums up to about $44$ TB. In view of the
very large scale training dataset, we adopted an asynchronous distributed parallel stochastic
gradient descent algorithm to speed up the training process nearly linearly on a
cluster of 16 GPUs.

\begin{figure}
  \vspace{-0.3cm}
  \includegraphics[width=0.5\textwidth]{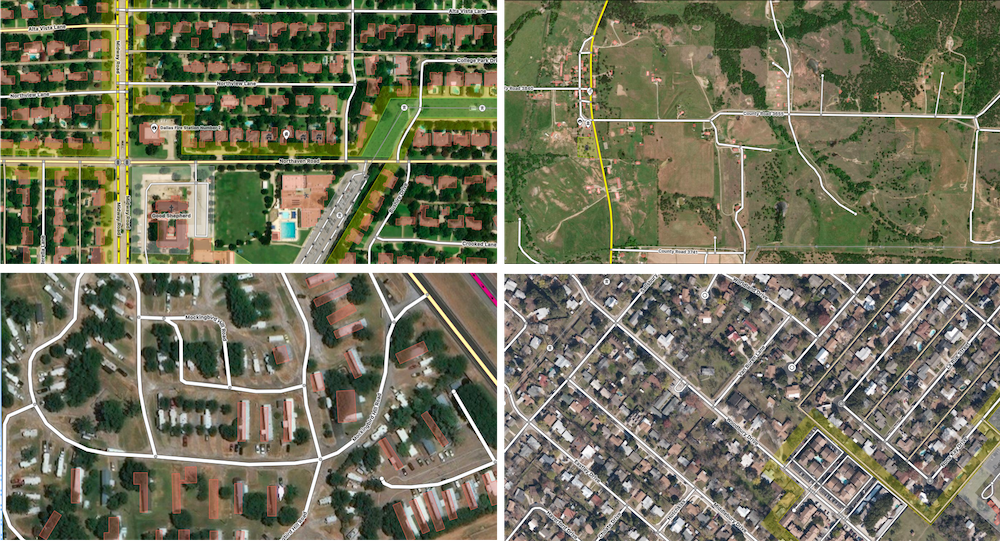}
  \caption{%
    Sample OSM data near Dallas, TX and San Antonio, TX: The plot on the upper
    left shows an urban area well labeled with most roads (white\slash yellow lines)
    and houses (red boxes) accurately marked. The sample on the upper right is
    drawn from a rural location with less accurate labels. The lower left plot represents
    another rural region with partially labeled houses. Finally, the lower right
    figure exemplifies a highly populated area without any house record in the OSM
    dataset, yet.
  }
    \vspace{-0.5cm}
  \label{fig:sample_OSM}
\end{figure}

We propose a training framework to incrementally generate maps from inaccurate and
incomplete OSM data with high resolution aerial images, using data
from four cities in Texas drawn from OSM and NAIP. Moreover, Las Vegas
data from SpaceNet have been employed as well. The salient contribution is the following:

\begin{itemize}
  \vspace{-0.1cm}
\item   We propose a training frame work to incrementally generate map from inaccurate
        and incomplete OSM maps.
\item   To our best knowledge, it is the first time OSM data is employed as a source
        of rasterized map imagery in order to train an pixel-to-pixel mapping from
        high resolution imagery.
\item   We analyzed the completeness of the OSM data and discuss an evaluation metric
        given the inaccuracy and incompleteness of the OSM labels.
\item   We numerically investigate the transfer of a model from one geo-location to
        another to quantify how well the models under consideration generalize.
\item   We tested an asynchronous parallel distributed stochastic gradient descent algorithm
        to speed up the training process nearly linearly on a cluster of 16 GPUs.
\item   Through a publicly accessible tool we showcase how we make available the
        our generated maps.
          \vspace{-0.1cm}
\end{itemize}

\section{Related Work}

Employing remote sensing imagery to generate maps has become more popular due to
the rapid progress in deep neural network in the past decade--particularly
in the arena of computer vision.
\begin{itemize}
   \vspace{-0.1cm}
\item   encoder-decoder--type convolutional neural networks for population estimation from
        house detection \cite{tiecke2017mapping}, or road detection with distortion tolerance
        \cite{zhou2018d}, or ensembles of map-information assisted U-Net models \cite{SpaceNetWinner2017}
\item   ResNet-like down-\slash upsampling for semantic segmentation of houses: \cite{RefineNetMicrosoft2018}
\item   image impainting for aerial imagery for semi-supervised segmentation:
        \cite{singh2018self}
\item   human infrastructure feature classification in overhead imagery with data prefiltering:
        \cite{bonafilia2019building}
           \vspace{-0.2cm}
\end{itemize}


\textit{Distributed} deep learning is the de-facto approach to accelerate its training. 
Until recently, it was believed that the asynchronous parameter-server-based distributed 
deep learning method is able to outperform synchronous distributed deep learning. 
However, researchers demonstrated synchronous training is superior to asynchronous 
parameter-server based training, both, from a theoretical and an empirical perspective
\cite{zhang2016staleness,GuptaZhang16,goyal2017accurate}. 
Nevertheless, the straggler problem remains a major issue in synchronous distributed training, 
in particular in a large scale setting. Decentralized deep learning \cite{NIPS2017_lian} is 
recently proposed to reduce latency issues in synchronous distributed training and researchers  
demonstrated that decentralized deep learning is guaranteed to have the same convergence 
rate as synchronous training. Asynchronous decentralized deep learning is further proposed 
to improve runtime performance while guaranteeing the same convergence rate as the 
synchronous approach \cite{LianZZL18}. Both scenarioes have been verified in theory
and practice over 100 GPUs on standard computer vision tasks such as ImageNet. 

\section{Data Corpus}

Since geo-spatial data comes in different geo-projections, spatial and temporal
resolution, it is critical to correctly reference it for consistent data preparation.
In this study, we used three sources of data: NAIP, OSM, and SpaceNet.
To obtain consistency in the generated training data, a tool developed by IBM was
utilized: \textit{PAIRS} \cite{klein2015pairs,lu2016ibm}, shorthand for Physical Analytics
Integrated data Repository and Services. The following sections describe in detail
how PAIRS processes each dataset to generate a uniform, easily accessible corpus 
of training data.


\subsection{Big Geo-Spatial Data Platform}
\label{sec:PAIRS}

\begin{figure}
   \vspace{-0.2cm}
  \includegraphics[width=0.4\textwidth]{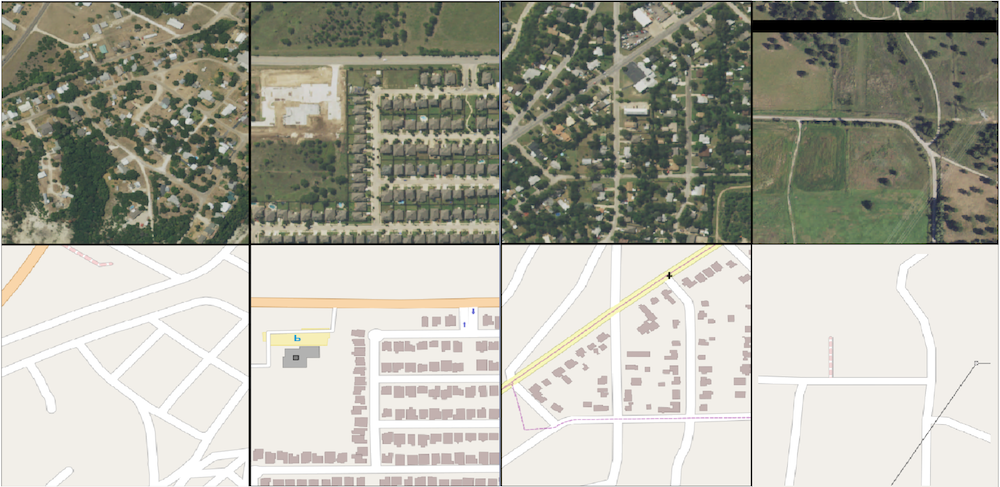}
  \caption{%
      Sample training data generated from PAIRS retrieval. The curation and geo-indexing
      of the PAIRS system aligns the aerial imagery and the rasterized OSM map to
      exactly match pixel by pixel.
  }
     \vspace{-0.5cm}
  \label{fig:sample_train}
\end{figure}

Both the NAIP imagery and OSM rasterized data are first loaded into IBM PAIRS.
PAIRS is a big spatio-temporal data platform that curates, indexes and manages a plurality
of raster and vector datasets for easy consumption by cross-datalayer queries and
geospatial analytics. Among others, datasets encompass remote sensing data including
satellite\slash aerial images, point cloud data like LiDAR (Light Detection And
Ranging) \cite{yan2015urban}, weather forecast data, sensor measurement data (cf.\
Internet of Things \cite{IoTDefIEEE2015}), geo-survey data, etc..

PAIRS \cite{ PAIRS_LU, klein2015pairs} is based on the scalable key-value store HBase\cite{vora2011hadoop}. For users
it masks various complex aspects of the geo-spatial domain where e.g.\ hundreds
of formats and geo-projections exist--provided by dozens of data sources. PAIRS
employs a uniform geo-projection across all datasets with nested indexing
(cf.\ QuadTree \cite{finkel1974quad}).
Raster data such as satellite images are cut into cells of size 32 by 32 pixels.
Each cell is stored in HBase indexed by a 16 bytes key encoding its spatial and
temporal information. Among others, the key design ensures efficient data storage
and fast data retrieval.


\subsection{Training Data Characteristics}
\label{sec:TrainDataCharacter}

USDA offers NAIP imagery products which are available either as digital ortho quarter
quad tiles or as compressed county mosaics \cite{USDANAIPDATA}. Each individual
image tile within the mosaic covers a $3.75$ by $3.75$ minute quarter quadrangle 
plus a $300$ meter buffer on all four sides. The imagery comes with 4 spectral bands
covering a red, green, blue, and near-infrared channel. The spatial resolution effectively
varies from half a meter up to about two meters. The survey is done over most of the
territory of the CONUS such that each location is revisited about every other year.

The open-source project OpenStreetMap (OSM) \cite{OSM} is a collaborative project
to create a freely available map of the world. The data of OSM is under tremendous
growth with over one million registered contributors\footnote{%
    as of February 2020, cf.\ \href{https://wiki.openstreetmap.org/wiki/Stats}{https://wiki.openstreetmap.org/wiki/Stats}
} editing and updating the map.
On the one hand, OSM data stem from GPS traces collected by voluntary field
surveys such as e.g.\ hiking and biking. On the other hand, a web-browser--based
graphical user interface with high resolution satellite imagery provides contributors
across the world an online tool to generate vector datasets annotating and updating
information on roads, buildings, land cover, points of interest, etc..

Both approaches have limitations. Non-military GPS is only accurate within meters--roads
labeled by GPS can be off by up to 20 meters. To the contrary, manual annotation
is time consuming and typically focused on densely populated areas like cities,
towns, etc.. Fig.\ \ref{fig:sample_OSM} shows examples of OSM labels with satellite
imagery as background for geo-spatial reference.
By visual inspection, it is evident that in most of the densely populated areas,
roads are relatively well labeled. Concerning houses the situation is similar.
However, compared to surrounding roads, more geo-spatial reference points need to
be inserted into the OSM database per unit area in order to label all buildings.
Thus, its coverage tends to lag behind road labeling. Shifting attention to rural
and remote areas, the situation becomes even worse, because there is less volunteers
available on site--or there is simply less attention and demand to build an accurate
map in such geographic regions.

Since our deeply-learnt models are conceptually based on a series of convolutional
neural networks to perform auto-encoding--type operations for the translation
of satellite imagery into maps, the problem under consideration could be rendered
in terms of image segmentation. Therefore, we pick a rasterized representation of
the OSM vector data, essentially projecting the information assembled by the OSM community
into an RGB-channel image employing the Mapnik framework \cite{MapnikGitRepo2005}.
This way each color picked for e.g.\ roads (white), main roads (yellow), highways
(blue), etc. and houses (brown) becomes a unique pixel segmentation label.

The rasterized OSM data is then geo-referenced and indexed in PAIRS such that for
each NAIP pixel there exists a corresponding unique label. The label generation
by the OSM rasterization is performed such that the OSM feature with the highest
z-order attribute\footnote{this index specifies an \textit{on-top} order} is picked.
E.g.\ if a highway crosses a local road by bridge, the highway color label is
selected for the RGB image rasterization. This procedure resembles the \textit{top-down}
view of the satellite capturing the NAIP imagery.
A notable technical aspect of PAIRS is its efficient storage: Besides data compression,
\textit{no-data} pixels (cf.\ the GeoTiff standard \cite{OCGGeoTiffStandard}) are
not explicitly stored. In terms of uploading the rasterized OSM data, this approach
significantly reduced the amount of disk space needed. Per definition, we declare
the OSM sandy background color as no-data.

After both NAIP and rasterized OSM data is successfully loaded into PAIRS, we use Apache
Spark \cite{zaharia2016apache} to export the data into tiles of $512 \times 512$
pixels. The uniform geo-indexing of the PAIRS system guarantees aerial imagery and
rasterized maps to exactly match at same resolution and same tile size. An illustration
of training samples shows Fig.\ \ref{fig:sample_train}. For this study, we focused
on the state of Texas, where we exported a total of about $2.8$ million images
with total volume of $2.6$ TB. However, we limit our investigation to the big cities
in Texas with rich human infrastructure, namely: Dallas, Austin, Houston, and San Antonio.


%

\subsection{SpaceNet: Data Evaluation Reference}

Since OSM data bear inaccuracies and it is incomplete in labeling, we picked
a reference dataset to test our approach against a baseline that has been well
established within the last couple of years within the remote sensing community.
\textit{SpaceNet} offers e.g.\ accurately labeled building polygons for the city
of Las Vegas, NV. Given a set of GeoJSON vector data files, PAIRS rasterized these
into the same style as employed by the OSM map discussed above. Then,
the corresponding result is curated and ingested as a separate raster layer, ready
to be retrieved from PAIRS the same way as illustrated in Fig.\ \ref{fig:sample_train}.
This setup allows us to apply the same training and evaluation procedure to be
discussed in the sections below.

\section{Models}

In this study we have used two existing deep neural network architectures with further
improvement on the baseline models in view of the characteristics of the data available
(cf.\ Sect.\ \ref{sec:intro}). Details of the models employed are described in the following.

\subsection{U-Net}

\textit{U-Net} \cite{Ronneberger_2015} is a convolutional neural network encoder-decoder
architecture. It is the state-of-the-art approach for image segmentation tasks.
In particular, U-Net is the winning algorithm in the 2nd round of the SpaceNet
\cite{etten2018spacenet} challenge on house detection using high resolution satellite
imagery. The U-Net architecture consists of a contracting path to capture context,
and employs a symmetric expanding path that enables precise localization. The contracting
path is composed of a series of convolution and max-pooling layers to coarse-grain
the spatial dimensions. The expanding path uses up-sampling layers or transposed
convolution layers to expand the spatial dimensions in order to generate a segmentation
map with same spatial resolution as the input image. Since the expanding path is
symmetric to the contracting path with skip connections wiring, the architecture
is termed \textit{U}-Net. Trainng the U-Net in a supervised manner for our remote
sensing scenario requires a pixel-by-pixel--matching rasterized map for each satellite
image.


\subsection{CycleGAN}

\begin{figure}
   \vspace{-0.2cm}
  \includegraphics[width=0.47\textwidth]{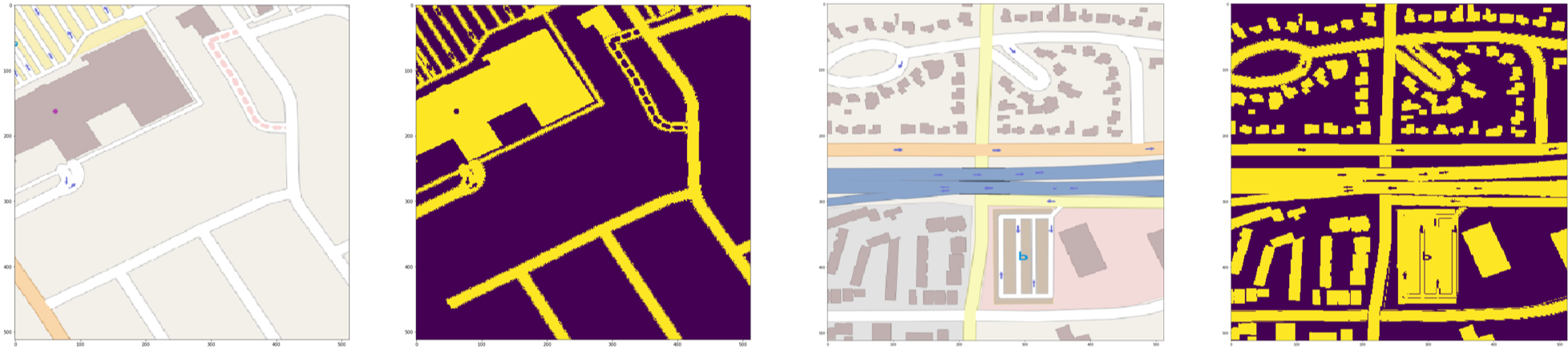}
     \vspace{-0.2cm}
    \caption{%
        Two pairs of samples of rasterized OSM maps (RGB color, left) with their
        corresponding feature mask (bi-color, right) next to each other.
    }
       \vspace{-0.5cm}
  \Description{}
  \label{fig:feature_extract}
\end{figure}

\begin{figure*}[t]
   \vspace{-0.2cm}
 \includegraphics[width=0.8\textwidth]{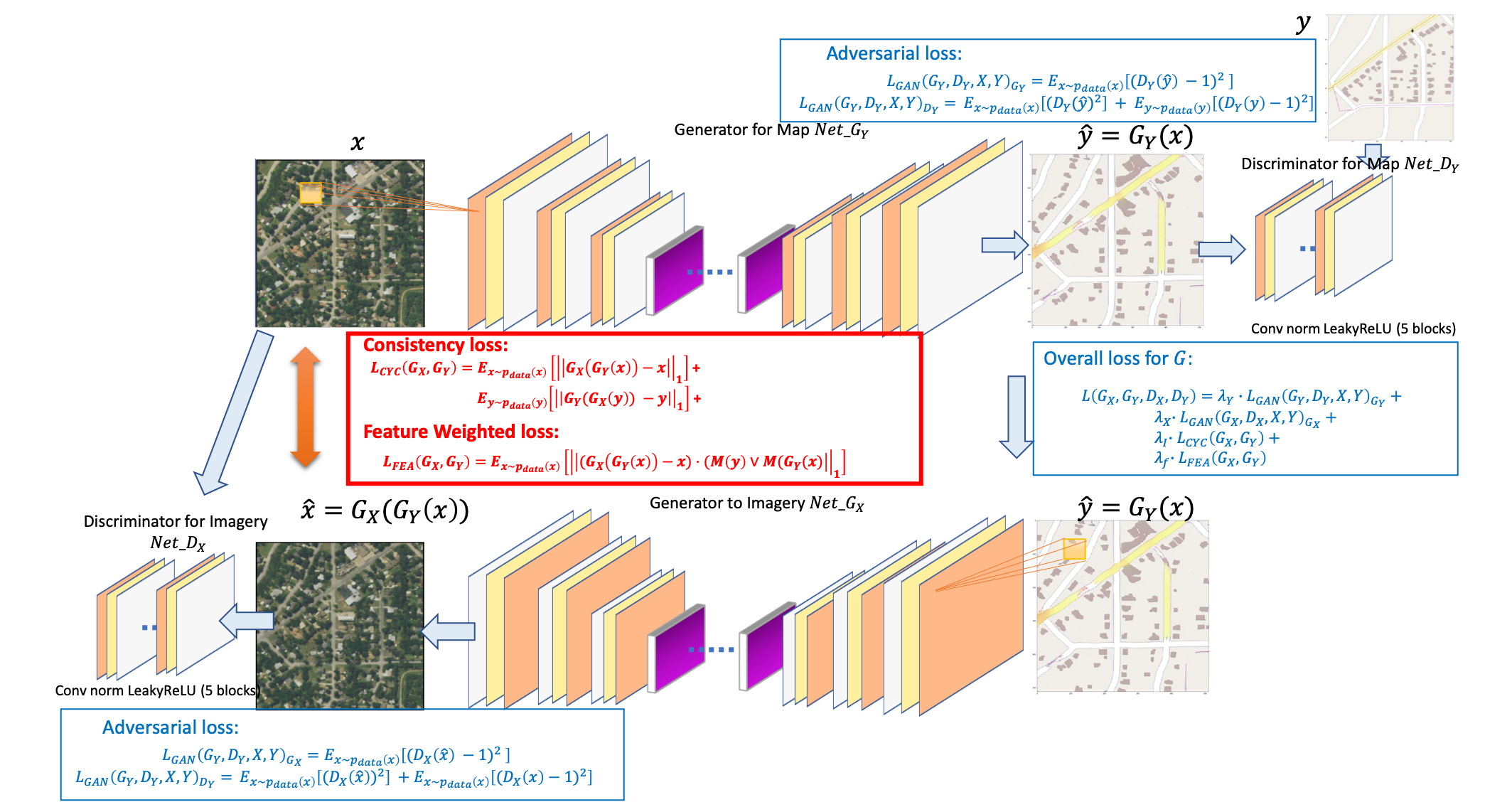}
 \caption{%
     Given satellite imagery data $x$ and corresponding map data $y$: illustration
     of our CycleGAN architecture showing the data flow from image input $x$, to generated
     map $\hat y$, to recreated image $\hat x=G_X(\hat y)$. This \textit{cycle}
     is used to compute a consistency loss $\lVert x-\hat x\rVert$ which is weighted
     by the feature map $M(y)$ yielding the FW-loss contribution during training.
 }
    \vspace{-0.5cm}
 \Description{}
  \label{fig:fwcg}
\end{figure*}
The $CycleGAN$ model was introduced in 2017 \cite{CycleGAN2017} for the task 
of image-to-image translation which is a class of computer vision problems with 
the goal to learn the mapping between two distributions of unpaired data sets
$X=\{x\}$ and $Y=\{y\}$. Given images $x$ from a source distribution $p_\text{data}(x)$
and maps $y$ from a target distribution $p_\text{data}(y)$, the task is to learn
a mapping $G:X \rightarrow Y$ such that the distribution of $G(x)$ is as close as
possible to the distribution of $p_\text{data}(y)$. In addition, a mapping
$F: Y \rightarrow X$ is established to further regulate the network's learning by
a so called \textit{cycle consistency loss} enforcing $F(G(x)) \approx x$. Starting
off with $y$ and repeating this line of reasoning, a second cycle consistency loss
pushes the $CycleGAN$'s numerical optimization towards $G(F(y))\approx y$.

The paper introducing $CycleGAN$ \cite{CycleGAN2017} provided a showcase that translated
satellite imagery to maps by way of example. Some rough, qualitative measurement on 
$CycleGAN$'s ability to convert overhead imagery to maps was provided. In our paper,
it is the first time $CycleGAN$ is evaluated quantitatively in terms of house and
road detection.

For the discussion to follow, $x\in X$ represents imagery input, $y\in Y$ map images.
The corresponding $CycleGAN$ generators are $\hat x=G_X(y)$ and $\hat y=G_Y(x)$, respectively.


\subsection{Feature-Weighted CycleGAN}
\label{sec:FWCycleGAN}

In generating maps from aerial images, the focus is to precisely extract well
defined features such as e.g.\ houses and roads from the geographic scene. Thus,
we added one more loss to the $CycleGAN$'s training procedure which we refer to as
\textit{feature-weighted} cycle consistency loss, \textit{FW loss} for short.
The FW loss is an attention mechanism putting more weight on pixel differences of
the cycle consistency loss which correspond to map features under consideration.
In the following section we describe in detail how features are defined and extracted,
and how the FW loss is computed.



As mentioned in Sect.\ \ref{sec:TrainDataCharacter}, for the rasterized OSM data,
houses and roads are labeled using a set of fixed colors $C$. For example,
houses (mostly residential) are labeled by a brownish color, cf.\ Fig.\
\ref{fig:feature_extract}. In contrast, roads can take a few colors such as plain
white, orange, light blue, etc..
We applied a color similarity-based method to extract the pixels that are detected
as features according to the predefined list of feature colors.
More specifically, the feature pixels from the OSM map $y$ and its
generated counterpart $\hat y=G_Y(x)$ is extracted by a color similarity function
$\delta$ generating a pixel-wise feature mask $M(y)$. $\delta$ is
defined by the international Commission on Illumination \cite{colorMetric}.
In particular, we used the formula referenced to as $CIE76$. The definition of the
binary feature mask $M(y)$ for a map $y$ is generated from a three-dimensional matrix
$y_{ijk}$ with indices $i,j$ representing gridded geospatial dimensions such as longitude
and latitude, and $k$ indexes the color channel of the map image:
   \vspace{-0.2cm}
\begin{equation}
M(y_{ijk}) =
\begin{cases}
    1,  & \delta(y_{ijk} ,C) \geq \Delta \\
    0,  & \text{else}
\end{cases}
\end{equation}
   \vspace{-0.0cm}
where $\Delta$ is a predefined threshold for the set of colors $C$.

Given $M(y)$, we added a FW loss $L_{FEA}$ to the generators' loss function in the
$CycleGAN$ model, defined by:
   \vspace{-0.1cm}
\begin{align}
L_{FEA} &(G_X, G_Y) = E_{x \sim P_{data}(x)} [\\\nonumber
        &\quad\lVert  (G_X(G_Y(x))-x  )\cdot( M(y) \vee M(G_Y(x) ) \rVert_{1}\\\nonumber
]       &
\end{align}
   \vspace{-0.3cm}



\section{Evaluation Metrics}

\begin{figure}

  \includegraphics[width=0.5\textwidth]{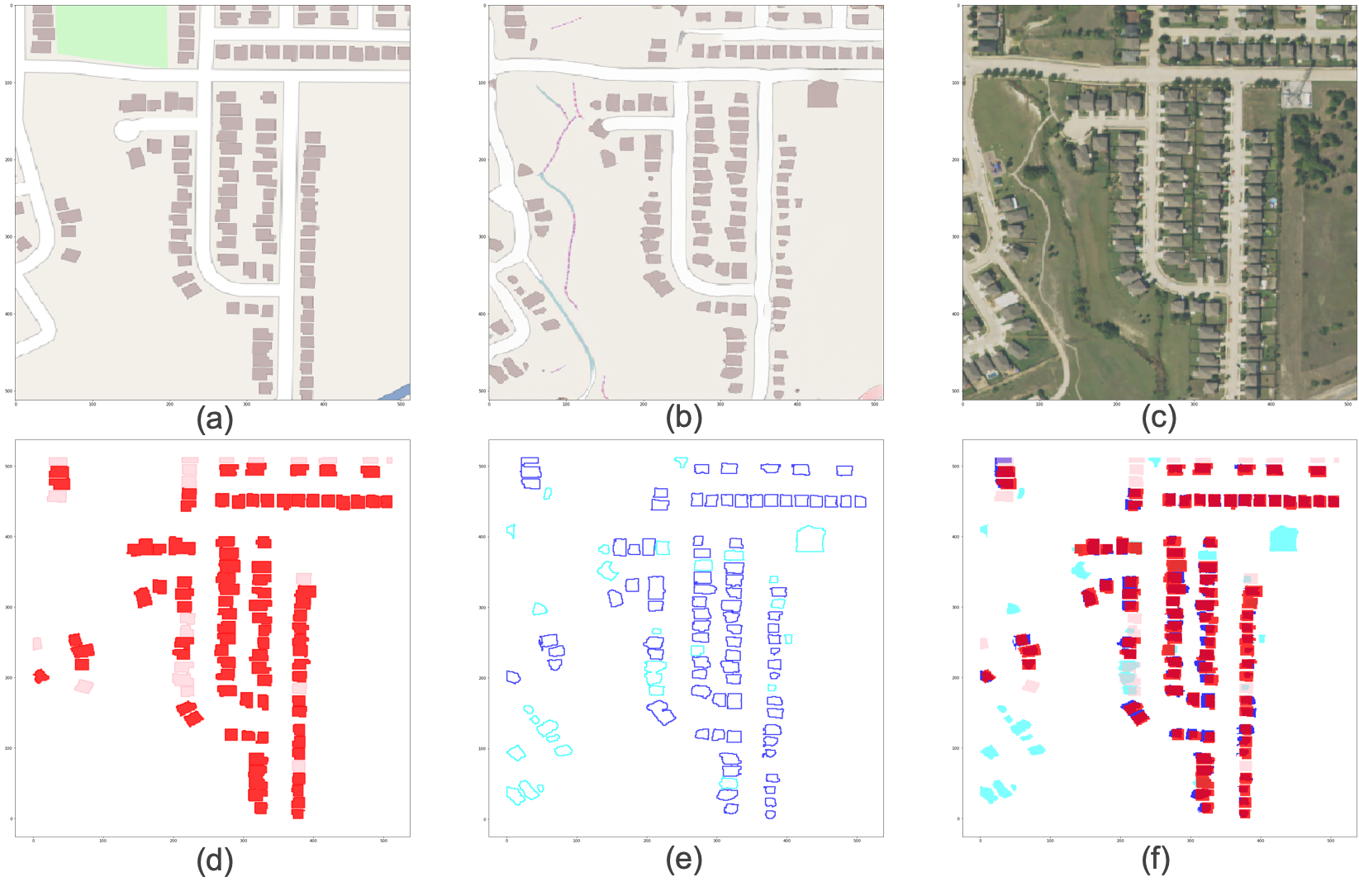}

  \caption{%
      Sample scenario from which to compute the $F1$ score: On the top row (a),
      (b) and (c) are the OSM map $y$, the generated map $\hat y$, and the NAIP image
      $x$, respectively. On the bottom row, (d), (e) and (f) are houses as labeled
      by the OSM map and houses detected. It results in the following cases colored:
      correctly detected houses (TP, red), missed houses (FN, pink), and false
      positive houses (FP, cyan).
  }
     \vspace{-0.5cm}
  \label{fig:f1}
\end{figure}

We adopted a feature-level detection score, similar to the SpaceNet evaluation approach:
Each house or road detected by the generated map $\hat y$ is evaluated against
the OSM map $y$ using a binary classification score $F1$ which consists of both,
$precision$ and $recall$. In the following section we detail on how each score
is computed.

In a first step, detected feature pixels in the maps (both, OSM map $y$ and generated
map $\hat y$) are extracted using the same method as described in Sect.\ \ref{sec:FWCycleGAN}.
Then a set of polygons $\{P^y_i\}$ and $\{P_j^{\hat y}\}$ is generated from the
extracted features. A feature like a house in the real map $y$ represented by a 
polygon $P_i^y$ is correctly detected if there exists a corresponding polygon
$P_j^{\hat y}$ in the generated map $\hat y$, such that the \textit{Intersection over Union}
(IoU)
\begin{equation}
    IoU_{ij}=\frac{
        \vert P_j^{\hat y}\cap P_i^y\vert
    }{
        \vert P_j^{\hat y}\cup P_i^y\vert
    }
\end{equation}
is greater than a given threshold $T$ where we used $T=0.3$ throughout our experiments.
The case counts as a \textit{true positive} ($tp_i=1$, $tp_i=0$ otherwise) for our 
evaluation. If there does not exist any $j$ that exceeds the IoU threshold $T$,
a \textit{false negative} ($fn_i=1$, $fn_i=0$ otherwise) is registered.
Vice versa, if there does not exist any $i$ of $\{P_i^y\}$, such that $IoU_{ij} \geq T$,
then the polygon $P_j^{\hat y}$ is counted as a \textit{false positive}
($fp_j=1$, $fp_j=0$ otherwise).

Fig.\ \ref{fig:f1} demonstrates examples of all three situations. The procedure
is repeated for all pairs of geo-referenced test data maps $y$ with corresponding
generated map $\hat y=G_Y(x)$.

The \textit{true positive} count of the data set $TP=\sum_{y,i} {tp_i}$ is the 
total number of \textit{true positives} for all samples $y \in Y$. 
In the same manner, the \textit{false positive} count is computed according to
$FP=\sum_{\hat y,i} {fp_i}$ from all $\hat y\in\{G_Y(x): x\in X\}$. Finally, we have
the \textit{false negative} count determined by $FN=\sum_{y,i} {fn_i}$.

Once the integrated quantities $TP$, $FP$, and $FN$ are obtained, precision $p$
and recall $r$ is computed by their standard definitions:
\begin{align}
    p &= TP/(TP+FP) \\
    r &= TP/(TP+FN)\quad.
\end{align}
In addition, the \textit{F1-score} that resembles the harmonic mean of precision
and recall is defined through
\begin{equation}
    f_1 = \frac{1}{\tfrac{1}{2}(1/r+1/p)}
        = 2\frac{pr}{p+r}\quad.
\end{equation}

As already discussed, neither is the OSM labels complete in terms of houses, nor
is it accurate in terms of roads for rural areas. By way of experiment, however,
we found that most of the house labels are accurate, if existing. Therefore, we
assume house labels to be incomplete, but accurate, and hence, we restrict ourself
to the recall score $r$ as a measure to evaluate model performance for detecting houses. 
We provide a discussion on the precision score $p$ as complement in Sect.\
\ref{sec:PrecisionDiscussion} employing human, visual inspection.


\section{Experimental Setup}

Our experiments were developed and performed on a cluster of 4 servers. Each machine
has 14-core Intel Xeon E5-2680 v4 2.40GHz processors, 1TB main memory, and 4 Nvidia 
P100 GPUs. GPUs and CPUs are connected via PCIe Gen3 bus with 16GB/s peak bandwidth
in each direction. The servers are connected by 100Gbit/s ethernet.

PyTorch version \texttt{1.1.0} is the underlying deep learning framework in use.
We use Nvidia's CUDA \texttt{9.2} API model, the CUDA-aware OpenMPI \texttt{v3.1.1},
and the GNU C++ compiler version \texttt{4.8.5} to build our communication library,
which connects with PyTorch via a Python-C interface.



\section{Results and Discussion}

As discussed above, in this study we focused on four cities in Texas, namely:
Austin, Dallas, San Antonio, and Houston. After all data tiles had been exported
from PAIRS, in a first step, we applied an entropy threshold on the tile's pixel
value distribution. It enabled us to filter out tiles dominated by bare land with
few features such as houses and roads. Then, each collection of tiles is randomly
split into training and testing with split ratio $4/1$.

It is found that among the four cities in Texas extracted for this study, the house
density\footnote{%
    defined as average number of houses labeled per square kilometer 
} varies significantly as summarized in Table \ref{tab:house_density}. 
Given the four cities are located in the same state, one would expect the density
of houses to be relatively similar, yet the number of house labels as provided by the OSM map
varies by more than one order of magnitude. 
For our setting, we consider the SpaceNet dataset as most complete in terms of house
labels. Although Las Vegas, NV is not in the state of Texas, lacking of a better alternative we used
the its house density for book-keeping purpose to compute the completeness score.
We define the completeness score by the ratio of the house density of any
Texas city vs.\ the house density in Las Vegas. The house densities and corresponding
completeness scores are listed in Table \ref{tab:house_density}.
House density is a critical variable for model performance, as a less complete
dataset generates a more biased model, thus impacting overall accuracy. After we present
our findings on model performance in view of data completeness below, we detail on
how we incrementally fill missing data in order to improve overall model accuracy.

\begin{table}[tbp]
\caption{%
    House density (average number of labeled houses per square kilometer) and
    completeness score for each dataset.
}
   \vspace{-0.2cm}
\scriptsize
\begin{tabular}{@{}lcc@{}}
\toprule
City        & House Density   & \multicolumn{1}{l}{Completeness Score} \\ \midrule
Vegas       & 3283          & 100\%                                  \\
Austin      & 1723          & 52\%                                   \\
Dallas      & 1285          & 39\%                                   \\
San Antonio & 95            & 3\%                                    \\
Houston     & 141           & 4\%                                    \\ \bottomrule
\end{tabular}
   \vspace{-0.2cm}
\label{tab:house_density}
\end{table}

\subsection{Model Comparison on Datasets with Different Level of Completeness}

In a first step, we established a comparison of \textit{U-Net} vs.\ \textit{CycleGAN}
using the most accurate and complete dataset from Las Vegas. Results are summarized
in Table \ref{tab:vegas}. As expected, the winning architecture in the SpaceNet
building detection challenge performs much better than the \textit{CycleGAN} model.
We note that for the SpaceNet Las Vegas dataset, houses are the only labels, i.e.\
in the rasterized map used for training, no road labels exist. In our experiments
we observed that CycleGAN is challenged by translating satellite images with rich
features such as roads, parks, lots, etc. into void on the generated map. Thus the
task is not necessarily suited for such an architecture.

\begin{table}[tbp]
   \vspace{-0.2cm}
\caption{U-Net and FW-CycleGAN Comparison on SpaceNet Vegas Dataset.}
   \vspace{-0.2cm}
\label{tab:vegas}
\scriptsize
\begin{tabular}{@{}llllll@{}}
\toprule
Model     & Train City & Test City & Precision & Recall & F1    \\ \midrule
U-Net     & Vegas      & Vegas     & 0.829     & 0.821  & \textbf{0.825} \\
CycleGAN & Vegas      & Vegas     & 0.700     & 0.414  & 0.520 \\ \bottomrule
\end{tabular}
   \vspace{-0.2cm}
\end{table}

\subsection{Model comparison on generalization}

In a next step, we wanted to investigate the generalization capability of the
\textit{U-Net} model trained on the accurate and complete SpaceNet data in Las Vegas.
If the model would be able to generalize from one geo-location to another,
the amount of data needed to train a model for the entire area of CONUS would be
significantly reduced. After we did train a \textit{U-Net} model on the SpaceNet Las Vegas
data, inference was performed on the Austin, TX dataset from OSM. 
The results are summarized in Table \ref{tab:unet_generalization}. We observe a
drop in recall from 82\% in Las Vegas, NV down to 25\% in Austin, TX.
Hence, the result underlines the need for a training dataset with wide variety 
of scenes across different geo-locations in order to be able to generate accurate maps.

\begin{table}[tbp]
\caption{The U-Net model generalization test cases.}
  \vspace{-0.2cm}
\label{tab:unet_generalization}
\scriptsize
\begin{tabular}{@{}llllll@{}}
\toprule
Model & Train City & Test City & Precision & Recall & F1    \\ \midrule
U-Net & Vegas      & Vegas     & 0.829     & 0.821  &\textbf{ 0.825 }\\
U-Net & Vegas      & Austin    & 0.246     & 0.253  & 0.250 \\ \bottomrule
\end{tabular}
  \vspace{-0.2cm}
\end{table}

Last, we compared the \textit{CycleGAN}, the \textit{FW-CycleGAN} and the \textit{U-Net}
models using the Austin dataset. Corresonpding results are shown in Table \ref{tab:three_model}.
We demonstrated that the additional FW loss significantly improved the recall of
the \textit{CycleGAN} increasing it to 74.0\% from a baseline $CycleGAN$ that yielded
a value of 46.4\%. Also, the \textit{FW-CycleGAN} model slightly outperformed the
\textit{U-Net} which achieved a recall of 73.2\%.

\begin{table}[tbp]
  \vspace{-0.2cm}
\caption{\textit{CycleGAN}, \textit{FW-CycleGAN} and \textit{U-Net} Comparison on PAIRS dataset.}
  \vspace{-0.2cm}
\label{tab:three_model}
\scriptsize
\begin{tabular}{@{}llllll@{}}
\toprule
Model        & Train City & Test City & Precision & Recall                                          & F1    \\ \midrule
CycleGAN    & Austin     & Austin        & 0.546     & 0.464                                           & 0.501 \\
FW-CycleGAN & Austin      & Austin    & 0.641    & \textbf{0.740}                               & 0.687 \\
U-Net        & Austin     & Austin            & 0.816     & 0.732                                           & 0.772 \\ \bottomrule
\end{tabular}
  \vspace{-0.2cm}
\end{table}

In yet another case study, we numerically determined the recall of the two best 
performing models, namely \textit{FW-CycleGAN} and \textit{U-Net}, on other cities.
The results are summarized in Table \ref{tab:generalization_four_city}.
As demonstrated, the \textit{FW-CycleGAN} model consistently generates better
recall values across all the three cities in Texas other than Austin. 

\begin{table}[tbp]
\caption{%
    Generalization comparison between \textit{FW-CycleGAN} and \textit{U-Net}
}
\label{tab:generalization_four_city}
  \vspace{-0.2cm}
\scriptsize
\begin{tabular}{@{}llllll@{}}
\toprule
Model          & Train City & Test City   & Precision & Recall & F1    \\ \midrule
FW-CycleGAN& Austin     & Austin      & 0.641    & \textbf{0.740}   & 0.687\\
U-Net          & Austin     & Austin      & 0.816     & 0.732         & 0.772 \\
FW-CycleGAN & Austin     & Dallas      & 0.495     & \textbf{0.626}  & 0.553 \\
U-Net          & Austin     & Dallas      & 0.512     & 0.498  & 0.505 \\
FW-CycleGAN & Austin     & San Antonio & 0.034     & \textbf{0.546}  & 0.063 \\
U-Net          & Austin     & San Antonio & 0.032     & 0.489  & 0.059 \\
FW-CycleGAN& Austin     & Houston     & 0.040     & \textbf{0.470}  & 0.074 \\
U-Net          & Austin     & Houston     & 0.045     & 0.370  & 0.080 \\ \bottomrule
\end{tabular}
  \vspace{-0.2cm}
\end{table}

\subsection{Incremental Data Augmentation}

Given the assumption that maps of cities from the same state follow the same
statistics regarding features, we propose a data augmentation scheme to incrementally
fill in missing labels in less complete datasets. The incremental data augmentation
scheme uses a model trained on a more completely labeled dataset, e.g., Austin area,
to generate maps for a less complete datasets, e.g., geo-locations of Dallas and San Antonio.

\begin{figure}
\vspace{-0.2cm}
  \includegraphics[width=0.5\textwidth]{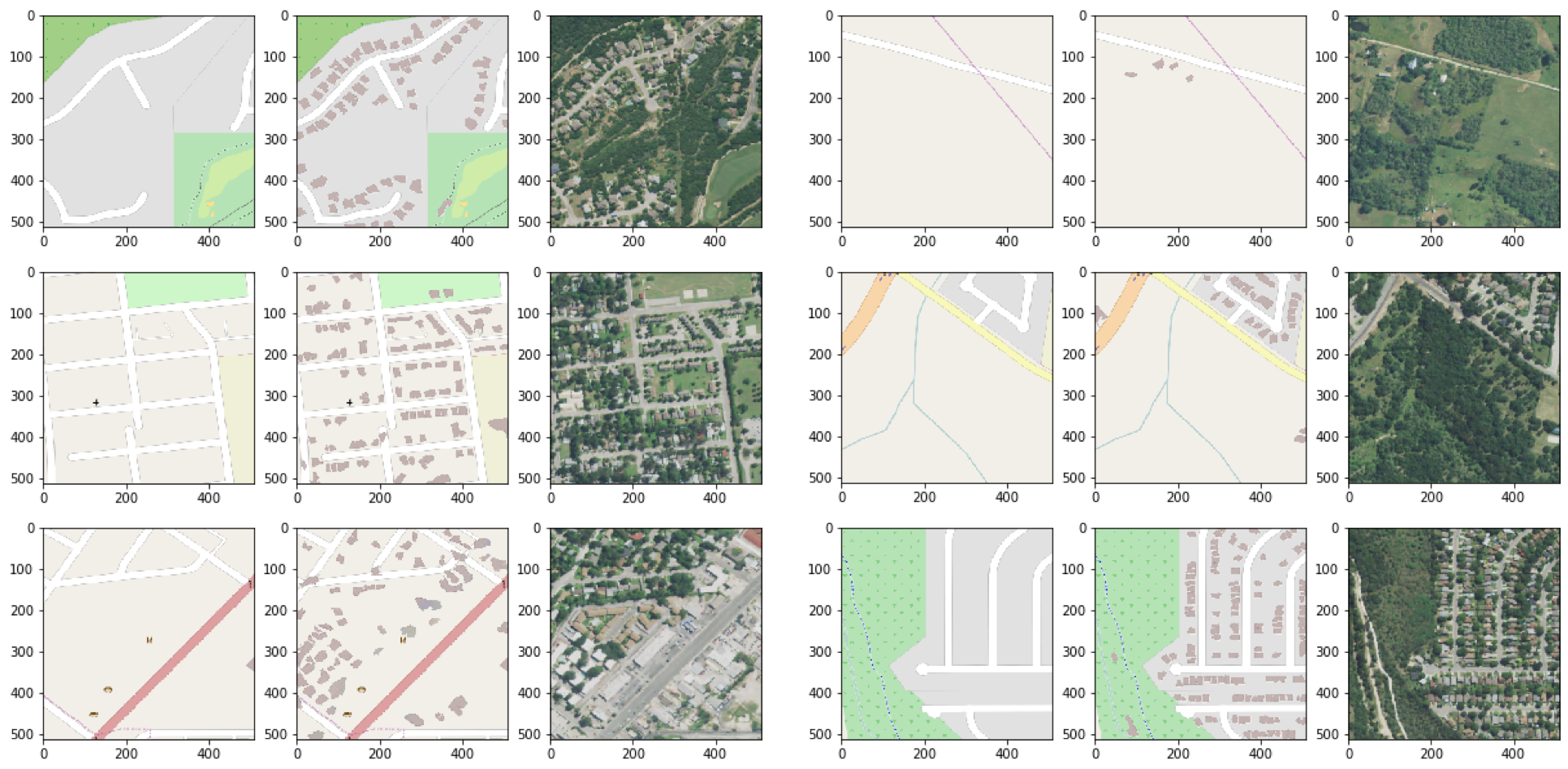}
  \caption{%
      Sample data augmentation, there are six samples in total. Three figures in each sample, thus there are two samples per row. The first figure in the sample is the original OSM map label, the second figure is the augmented figure, and the last one is the high resolution imagery as a reference to show the validity of the augmented map labels. As is shown, only false positive houses in the generated map is added back the to the original OSM map to generate the augmented map.  
  }
  \vspace{-0.5cm}
  \Description{}
  \label{fig:data_aug}
\end{figure}

More specifically, due to consistently better generalization score shown in Table \ref{tab:generalization_four_city}
we used the \textit{FW-CycleGAN} model trained on Austin data 
to generate augmented maps. The houses labeled as false positive in the generated
maps are added to the original maps to create the augmented maps $y\prime\in Y$. 
In this study, we also generated the augmented map for Austin. A sample of the augmented 
map compared to the original map as well as corresonding satellite imagery is 
shown in Fig. \ref{fig:data_aug}. 

By data augmentation, the average house density of Austin increased by about 25\%
to $2168$ per square kilometer. The house density of Dallas and San Antonio has
been increased to a level close to Austin. Numerical details provided in Table \ref{tab:hd_new}.
It is noted that data augmentation is only performed on the training dataset,
the testing dataset remains the same throughout the experiment.

\begin{table}[]
  \vspace{-0.2cm}
\caption{House Density comparison with data augmentation.}
  \vspace{-0.2cm}
\label{tab:hd_new}
\scriptsize
\begin{tabular}{@{}lll@{}}
\toprule
City     & OSM  & Augmented \\
        & House Density & House Density \\ \midrule
Austin      & 1723                   & 2168                      \\
Dalla       & 1285                   & 1678                    \\
San Antonio & 95                     & 1259                    \\\bottomrule
\end{tabular}
  \vspace{-0.2cm}
\end{table}

The model accuracy from both, the original OSM map $y$ and its augmented counterpart
$y\prime$ for the three cities under consideration are shown in Table \ref{tab:model_aug}.
As obvious, the models trained using the augmented map outperform the models trained
using original OSM, in particular for cities with less labels in OSM. For example,
the recall score got lifted from 11.8\% to 51.4\% for the \textit{U-Net} in the
San Antonio area. Even for Austin, where the data are most complete, recall improved
from 73.2\% to 84.9\% which is almost close to the SpaceNet winning solution,
which has a F1 score of 88.5\% for Las Vegas. We note that, compared to our training data
the SpaceNet winning solution was achieved by using an 8-band multispectral
data set plus OSM data trained on a more completely and accurately labeled dataset.
It is also noted that we do employ
a smaller IoU threshold than the SpaceNet metrics. Nevertheless, 
there are certainly more trees in Texas, which highly impact the detected area of a house. Tree cover is
the main reason we reduce the IoU threshold to a smaller value of 0.3.


\begin{table}[]
\caption{Model accuracy comparison using original OSM map and augmented map. }
\vspace{-0.2cm}
\label{tab:model_aug}
\scriptsize

\begin{tabular}{@{}llllllll@{}}
\toprule
city                         & model        & \multicolumn{2}{l}{Precision} & \multicolumn{2}{l}{Recall} & \multicolumn{2}{l}{F1}  \\ \midrule
                             &              & OSM   & Augmented & OSM   & Augmented & OSM   & Augmented \\ \cmidrule(l){3-8} 
\multirow{2}{*}{Austin}      & FW-Cycle-Gan & 0.641 & 0.614     & 0.74  & \textbf{0.769 }    & 0.687 & 0.683     \\
                             & U-Net        & 0.816 & 0.7       & 0.732 & \textbf{0.849}     & 0.772 & 0.768     \\
\multirow{2}{*}{Dallas}      & FW-Cycle-Gan & 0.524 & 0.536     & 0.51  & \textbf{0.761}     & 0.517 & 0.629     \\
                             & U-Net        & 0.765 & 0.633     & 0.772 & \textbf{0.830}      & 0.768 & 0.718     \\
\multirow{2}{*}{San Antonio} & FW-Cycle-Gan & 0.082 &  0.020         & 0.133 &     \textbf{0.409}      & 0.101 &     0.039      \\
                             & U-Net        & 0.179 & 0.026     & 0.118 & \textbf{0.514}     & 0.142 & 0.049     \\ \bottomrule
\end{tabular}
\vspace{-0.2cm}
\end{table}

\subsection{Discussion on Precision}
\label{sec:PrecisionDiscussion}

The precision score evaluated using OSM labels as ground truth is negatively
impacted by the incompleteness of the OSM data. Indeed, we observed that OSM labels
frequently miss houses whose presence is evident from NAIP imagery. 
In order to provide a quantitative understanding on the true model performance, 
we took the Austin test dataset as a case study: We randomly picked 100 samples
of the aerial images, and manually surveyed these images to correct the incomplete
labels of OSM. Using the corrected labels as the ground truth, $TP$, $FN$, and
$FP$ were computed for the subset.

Corresponding results are shown in Table \ref{tab:manual_count}. Comparing Table
\ref{tab:manual_count} with Table \ref{tab:three_model}, the F1 score of both
\textit{U-Net} and \textit{FW-CycleGAN} improved with the manual label correction.
While the recall scores are the same, the \textit{U-Net} yields a higher precision score, 
thus resulting in a higher F1 score of 87.5\%. The improved score is indeed expected,
as many false positives turned into true positives after the correction of incomplete
OSM labels. Moreover, the models trained with incomplete labels will likely under-estimate
the positives, thus both resulted in a higher precision and lower recall score. 
Overall, \textit{U-Net} out-performs \textit{FW-CycleGAN} in this relatively more
accurate dataset among four cities under study. 

Remarkably, this limited preliminary work showed that after data augmentation,
in spite of the fact that the \textit{U-Net} is trained using inaccurate and incomplete
OSM training labels, its F1 score performance is par to the winning SpaceNet solution
which was trained and validated on accurate datasets. we plan to perform more
comprehensive manual checking on other dataset in the future.

\begin{table}[tbp]
\caption{Corrected scores for \textit{FW-CycleGAN} and \textit{U-Net} with manual count on Austin dataset.}
\vspace{-0.2cm}
\label{tab:manual_count}
\scriptsize
\begin{tabular}{@{}lllllll@{}}
\toprule
Model       & TP   & FP  & FN  & Precision & Recall & F1                              \\ \midrule
FW-CycleGAN & 4112 & 758 & 956 & 0.844     & 0.811  & 0.828                           \\
U-Net       & 3817 & 203 & 889 & 0.950     & 0.811  & \textbf{0.875} \\ \bottomrule
\end{tabular}
\vspace{-0.5cm}
\end{table}

\subsection{Distributed training experiment}

We used Decentralized Parallel SGD (DPSGD) \cite{NIPS2017_lian} to accelerate training. 
We rearranged the weight update steps for both the generators and the discriminators
such that the stepping function of the stochastic gradient descent algorithm renders
simutaneously. This way, the weight updating and averaging step of the two sets
of networks (generators and discriminators) remains the same 
 as an architecture that employs one network and one objective function, which is 
originally proposed in \cite{NIPS2017_lian}. All learners (GPUs) are placed in a communication ring. 
After every mini-batch update, each learner randomly picks another learner in the
communication ring to perform weight averaging, as proposed in \cite{icassp20}.
In addition, we overlap the gradient computations by the weight communication and
averaging step to further improve runtime performance. As a result, we achieved a
speed-up of 14.7 utilizing 16 GPUs, hence reducing the training time of \textit{CycleGAN} 
from roughly 122 hours on one GPU down to 8.28 hours employing the decentralized parallel
training utilizing 16 GPUs.


\section{Public Access to Generated Maps}

Given the big geo-spatial data platform PAIRS discussed in Sect.\ \ref{sec:PAIRS},
we make available inferred maps of our models to the public. The open-source Python
module \texttt{ibmpairs} can be downloaded from
\href{https://github.com/ibm/ibmpairs}{https://github.com/ibm/ibmpairs}\footnote{%
    Corresponding \texttt{pip} and \texttt{conda} package are available through
    \href{https://pypi.org/project/ibmpairs/}{https://pypi.org/project/ibmpairs/}
    and \href{https://anaconda.org/conda-forge/ibmpairs}{https://anaconda.org/conda-forge/ibmpairs},
    respectively, i.e. running \texttt{pip install ibmpairs} or
    \texttt{conda install -c conda-forge ibmpairs} is a way to get the Python module.
}. In order to retrieve the generated map features 
as
colored overlay along with geo-referenced NAIP RGB image in the background,
the following JSON load
\vspace{-0.1cm}
\begin{lstlisting}[basicstyle=\scriptsize, language=Python]
{ "layers": [
    { "id": "50155"}, {"id": "50156"}, {"id": "50157"},
    { "id": "49238", "aggregation": "Max",
      "temporal":
      {"intervals": [{"start": "2014-1-1","end": "2016-1-1"}]}
    },
    { "id": "49239", "aggregation": "Max",
      "temporal":
      {"intervals": [{"start": "2014-1-1","end": "2016-1-1"}]}
    },
    { "id": "49240", "aggregation": "Max",
      "temporal":
      {"intervals": [{"start": "2014-1-1","end": "2016-1-1"}]}
    }
  ],
  "temporal": {"intervals": [{"snapshot": "2019-4-16"}]},
  "spatial": {
    "coordinates": [32.6659568,-97.4756499, 32.6790701,-97.4465533],
    "type": "square"
  }
}
\end{lstlisting}
\vspace{-0.1cm}
can be submitted as query to PAIRS. The Python sub-module \texttt{paw} of \texttt{ibmpairs}
utilizes the PAIRS's core query RESTful API. Example code reads:
\vspace{-0.1cm}
\begin{lstlisting}[basicstyle=\scriptsize, language=Python]
# set up connection to PAIRS
from ibmpairs import paw
import os
os.environ['PAW_PAIRS_DEFAULT_PASSWORD_FILE_NAME'] = \
    '/path/to/pairs/password/file'
os.environ['PAW_PAIRS_DEFAULT_USER'] = \
    'pairsuser@somedomain.org'
paw.load_environment_variables()
# retrieve generated map and associated imagery from PAIRS
query = paw.PAIRSQuery(queryJSON)
query.submit()
query.poll_till_finished()
query.download()
query.create_layers()
\end{lstlisting}
\vspace{-0.1cm}
with \texttt{queryJSON} the JSON load listed above, assuming a fictitious user
\texttt{pairsuser@somedomain.org}. Overall, six PAIRS raster layers are queried
corresponding to 3 RGB channels for the satellite imagery and 3 RGB channels for
the generated, rasterized map.

\section{Conclusion and Future Work}

In this paper, we performed a case study to investigate the quality of publicly
available data in the context of map generation from high-resolution aerial/satellite
imagery by applying deep learning architectures. In particular, we utilized
the aerial images from the NAIP program to characterize rasterized maps based on
the crowdsourced OSM dataset. 

We confirmed that geographic, economic, and cultural heterogeneity renders significant 
differences on man-made infrastructure which calls for more broadly available
training data like the ones used in this study.
We employed two state-of-the-art deep convolution neural network models, namely
\textit{U-Net} and \textit{CycleGAN}. Furthermore, based on the objective, we 
introduced the \textit{Feature-Weighted CycleGAN} which significantly improved 
binary classification accuracy for house detection. Although OSM is not accurate 
in rural areas, and it is incomplete in urban areas, we assumed: once a house is
labeled, it is accurately labeled. Consequently, we focused our model performance
evaluation on recall. In addition, we provided manually obtained evaluation metrics, 
to show that both precision and recall increases in value in accurately labeled subsets.

For scenarios where the incompleteness of OSM labels is significant, we propose
an incremental data augmentation scheme that has significantly improved model accuracy
in such areas. Even for cities which are relatively complete in terms of labeling,
the data augmentation scheme helped lifting the best recall to 84\%, and our manual count 
 of binary classification of the Austin dataset shows the precision score is above 84\%, yielding a F1 score prominently
close to a corresponding SpaceNet winning solution exploiting more data input compared
to our approach.

Obviously, to finally map the entire world we need to deal with enormous amounts
of training data. To this end, we applied an Decentralized Parallel Stochastic 
Gradient Descent (DP-SGD) training scheme that is scalable to hundreds of GPUs
with near linear speed-up. At the same time it carries the same level of convergence
compared to non-parallel training schemes. We demonstrated an implementation of
the DP-SGD scheme and achieved a speed up of $14.7$ times over a cluster of 16 GPUs.
Nevertheless, we observed a gap in convergence. Tuning the code for further improvement
is the subject of our current, ongoing agenda.


\bibliographystyle{ACM-Reference-Format}
\bibliography{image_map_ref}

\appendix

\end{document}